# Partial-Order Planning with Concurrent Interacting Actions

**Craig Boutilier**                                                    CEBLY@CS.TORONTO.EDU
*Department of Computer Science*
*University of Toronto*
*Toronto, ON, M5S 3H8, Canada*

**Ronen I. Brafman**                                                   BRAFMAN@CS.BGU.AC.IL
*Department of Computer Science*
*Ben-Gurion University*
*Beer Sheva, Israel 84105*

## Abstract

In order to generate plans for agents with multiple actuators, agent teams, or distributed controllers, we must be able to represent and plan using concurrent actions with interacting effects. This has historically been considered a challenging task requiring a temporal planner with the ability to reason explicitly about time. We show that with simple modifications, the STRIPS action representation language can be used to represent interacting actions. Moreover, algorithms for partial-order planning require only small modifications in order to be applied in such multiagent domains. We demonstrate this fact by developing a sound and complete partial-order planner for planning with concurrent interacting actions, POMP, that extends existing partial-order planners in a straightforward way. These results open the way to the use of partial-order planners for the centralized control of cooperative multiagent systems.

## 1. Introduction

In order to construct plans for agents with multiple actuators (such as multi-armed robots), agent teams, or controllers distributed throughout an environment, we must be able to model the effects and interactions of multiple actions executed concurrently, and generate plans that take these interactions into account. A viable solution to the basic multiagent/multi-actuator planning (MAP) problem must include economical action descriptions that are convenient to specify and are easily manipulable by planning algorithms, as well as planning methods that can deal with the interactions generally associated with concurrent actions.

Surprisingly, despite the interest in multiagent applications—for instance, in robotics (Donald, Jennings, & Rus, 1993; Khatib, Yokoi, Chang, Ruspini, Holmberg, Casal, & Baader, 1996) and distributed AI (e.g., see the various proceedings of the International Conference on Multiagent Systems)—and the large body of work on distributed multiagent planning, very little research addresses this basic problem of planning in the context of concurrent interacting actions. Researchers in distributed AI have considered many central issues in multiagent planning and multiagent interaction, but much existing research is concerned mainly with problems stemming from the distributed nature of such systems, such as task decomposition and resource allocation (Durfee & Lesser, 1989; Wilkins & Myers, 1998; Stone & Veloso, 1999), obtaining local plans that combine to form global plans (Durfee & Lesser, 1991; Ephrati, Pollack, & Rosenschein, 1995), minimizing communication needs (Wolverton & des Jardins, 1998; Donald et al., 1993), and so on. As opposed to this form of *distributed* planning, our focus in this paper is on *centralized* planning for agent teams (or distributed actuators).

Representation of concurrent actions has been dealt with by various researchers in the knowledge-representation community (e.g., Lin & Shoham, 1992; Reiter, 1996; de Giacomo, Lésperance, & Levesque, 1997; Moses & Tennenholtz, 1995; Pinto, 1998). Of particular note are the action languages $A_c$ (Baral & Gelfond, 1997) and $C$ (Giunchiglia & Lifschitz, 1998) which enable the specification of concurrent interacting actions and employ a nonmonotonic override mechanism to deduce the effects of a set of actions with conflicting effects. Finally, a number of contemporary planners can handle concurrent *noninteracting* actions to a certain degree—examples include Graphplan (Blum & Furst, 1995), and IPP (Koehler, 1998), which extends Graphplan to handle resource constraints, and more recently OBDD-based planners such as MBP (Cimatti, Giunchiglia, Giunchiglia, & Traverso, 1997) and UMOP (Jensen & Veloso, 2000)—while Knoblock (1994) provides a good discussion of the issue of parallelizing serial plans.

Despite these advances, one often sees in the planning community suggestions that *temporal planners* are required to adequately deal with concurrent interacting actions. For example, in his discussion of parallel execution plans, Knoblock (1994) asserts:

> To handle these cases [of interacting actions] requires the introduction of an explicit representation of time, such as that provided in temporal planning systems.

A similar perspective seems implicit in the work on parallel action execution presented by Lingard and Richards (1998). Certainly time plays a role in planning—in any planner the idea that sequences of actions occur embodies an implicit notion of time. However, we disagree that time in centralized multiagent planning must be dealt with in a more explicit fashion than in single-agent planning. The main aim of this paper is to demonstrate that the MAP problem can be solved using very simple extensions to existing (single-agent) planners like UCPOP (Penberthy & Weld, 1992). We provide a representation and MAP algorithm that requires no explicit representation of time. This is not to deny that explicit temporal representations are useful in planning—for many problems these may be necessary—but we do not think this is the key bottleneck in planning the activities of multiagent teams. Specifically, we view temporal issues to be orthogonal to the main concerns facing multiagent planning.

The central issue in multiagent planning lies in the fact that individual agent actions *do* interact. Sometimes planning is hindered as a result of action interaction: action $X$ of agent 1 might destroy the intended effect of action $Y$ of agent 2 if executed concurrently. For example, in a half-duplex communication line, we cannot allow simultaneous transmission of messages from both sides. In such a case, a planning algorithm has to make sure that $X$ and $Y$ are not executed at the same time. More interesting is the fact that planning often *benefits* as a result of action interaction: action $X$ of agent 1 might only achieve an intended effect if agent 2 performs action $Y$ concurrently. For example, opening a typical door requires two simultaneous actions: turning the knob and pushing the door. In military activities, different units may have to coordinate their actions in order to be effective (e.g., turn on engines or lights simultaneously, or attack at the same time). Similar situations arise in a variety of domains. In such cases, a planning algorithm has to ensure that the appropriate actions are executed at the same time. An action representation that makes these interactions explicit and a planning algorithm that can, as result of these interactions, prescribe that certain actions must or must not be executed concurrently are some of the main features of any multiagent planner. Temporal representations may play a role in the scheduling of such actions, but are not strictly necessary for reasoning about the effects of interaction (or lack thereof).

To illustrate some of these issues, consider the following example which will be discussed in detail later in the paper: two agents must move a large set of blocks from one room to another. While they could pick up each block separately, a better solution would be to use an existing table in the following manner. First, the agents put all blocks on the table, then they each lift one side of the table. However, they must lift the table simultaneously; otherwise, if only one side of the table is lifted, all the blocks will fall off. Having lifted the table, they must move it to the other room. There they put the table down. In fact, depending on the precise goal and effects of actions, it may be better for one agent to drop its side of the table, causing all of the blocks to slide off at once. Notice how generating this plan requires the agents to coordinate in two different ways: first, they must lift the table together so that the blocks do not fall; later, one of them (and only one) must drop its side of the table to let the blocks fall.

Since the actions of distinct agents interact, we cannot, in general, specify the effects of an individual's actions without taking into account what other actions might be performed by other agents at the same time. That *truly* concurrent actions are often desirable precludes the oft-used trick of "interleaving semantics" (Reiter, 1996; de Giacomo et al., 1997). Agents lifting a table on which there are a number of items must do so simultaneously or risk the items sliding from the table, perhaps causing damage. Interleaving individual "lift my side of table" actions will not do.

One way to handle action interactions is to specify the effects of all *joint actions* directly. More specifically, let $A_i$ be the set of actions available to agent $i$ (assuming $n$ agents labeled $1 \ldots n$), and let the *joint action space* be $A_1 \times A_2 \times \cdots \times A_n$. We treat each element of this space as a separate action, and specify its effects using our favorite action representation.[1]

The main advantage of this reduction scheme is that the resulting planning problem can be tackled using any standard planning algorithm. However, it has some serious drawbacks with respect to ease of representation. First, the number of joint actions increases exponentially with the number of agents. This has severe implications for the specification and planning process. Second, this reduction fails to exploit the fact that a substantial fraction of the individual actions may not interact at all, or at least not interact under certain conditions. We would like a representation of actions in multiagent/multi-actuator settings that exploits the independence of individual action effects to whatever extent possible. For instance, while the lift actions of the two agents may interact, many other actions will not (e.g., one agent lifting the table and another picking up a block). Hence, we do not need to explicitly consider all combinations of these actions, and can specify certain individual effects separately, combining the effects "as needed."

---

1. Our discussion will center on the **STRIPS** action representation, but similar considerations apply to other representations such as the situation calculus (McCarthy & Hayes, 1969; Reiter, 1991) and dynamic Bayes nets (Dean & Kanazawa, 1989; Boutilier & Goldszmidt, 1996).

Joint actions also cause problems for the planning process itself: their use in the context of most planners forces what seems to be an excessive degree of commitment. Whenever the individual action of some agent can accomplish a desired effect, we must insert into our plan a joint action, thereby committing all other agents to *specific* actions to be executed concurrently, even though the actual choices may be irrelevant. For these reasons, we desire a more "distributed" representation of actions.

We are therefore faced with the following two problems:

1. The representation problem: how do we naturally and concisely represent interactions among concurrently executed actions.

2. The planning problem: how do we plan in the context of such a representation.

In this paper, we show how the STRIPS action representation can be augmented to handle concurrent interacting actions and how existing nonlinear planners can be adapted to handle such actions. In fact, it might come as a surprise that solving both problems requires only a small number of changes to existing nonlinear planners, such as UCPOP (Penberthy & Weld, 1992).[2] The main addition to the STRIPS representation for action $a$ is a *concurrent action list*: this describes restrictions on the actions that can (or cannot) be executed concurrently in order for $a$ to have the specified effect (indeed, $a$ can have a number of different *conditional effects* depending on which concurrent actions are applied). In order to handle this richer language, we must make a number of modifications to "standard" partial-order planners: (a) we add equality (respectively, inequality) constraints on action orderings to enforce concurrency (respectively, nonconcurrency) constraints; and (b) we expand the definition of *threats* to cover concurrent actions that could prevent an intended action effect.

We emphasize that we deal with the problem of planning the activities of multiple agents or agents with multiple actuators in a centralized fashion, as opposed to *distributed* planning. Our model assumes that one has available a central controller that can decide on an appropriate joint plan and communicate this plan to individual agents (or actuators). While distributed planning is an important and difficult problem, it is not the problem addressed in this work. We also assume that some mechanism is available by which individual agents can ensure that the *execution* of their concurrent plans are synchronized. Again, while an issue of significance and subtlety, it is not a task we consider in this paper.

We note that planning with parallel actions has been addressed in some detail by Lingard and Richards (1998). Specifically, they provide a very general framework for understanding constraint-posting, least-commitment planners that allow for concurrent action execution. However, as mentioned above, their work takes an explicit temporal view of the problem and focuses primarily on issues having to do with action duration. Furthermore, while multiagent planning could presumably be made to fit within their model, this seems not to be their main motivation. In fact, the planning algorithms they discuss deal with the issue of ensuring that parallel actions do not have *negative* synergistic effects, and explicitly exclude the possibility of positive synergy. In our work, we abstract away from the temporal component and focus precisely on planning in the presence of such synergies, both positive and negative.

---

2. Moreover, other planning algorithms (e.g., Blum & Furst, 1995; Kautz & Selman, 1996) should prove amenable to extension to planning with concurrent interacting actions using similar ideas.

In the following section we describe our STRIPS-style representation for concurrent, interacting actions and multiagent plans. In Section 3 we describe the *Partial-Order Multiagent Planning* algorithm (POMP), a modified version of the UCPOP algorithm that can be used to generate plans for multiagent teams or multiactuator devices. Section 4 illustrates the POMP algorithm on an extended example. In Section 5 we discuss the soundness and completeness of the POMP algorithm. We conclude in Section 6 with a discussion of some issues raised by this work.

## 2. Representing Concurrent Actions and Plans

We begin by considering the representation of concurrent actions and partially ordered plans using a simple extension of traditional planning representations. We first describe a standard action representation based on the STRIPS model, specifically that used by UCPOP (Penberthy & Weld, 1992). We then describe the extension of this representation to represent concurrent interacting actions and its semantics, and finally describe the representation and semantics of partially ordered multiagent plans.

### 2.1 The STRIPS Action Representation

Variants of the STRIPS action representation language (Fikes & Nilsson, 1971) have been employed in many planning systems. We assume a finite set of predicates and domain objects (generally typed) that characterize the domain in question. *States* of this system are truth assignments to ground atomic formulae of this language. A state is represented as a set (or conjunction) of those ground atoms true in that state, such as

$$\{OnTable(B1), Holding(A, B2)\}$$

thus embodying the *closed world assumption* (Reiter, 1978). *Actions* induce state transitions and can be viewed as partial mappings from states to states. An action $A$ is represented using a *precondition* and an *effect*, each a conjunction of literals (sometimes referred to as the precondition or effect *list*). If a state does not satisfy the conjunction of literals in the precondition list, the effect of applying the action is undefined. Otherwise, the state resulting from performing action $A$ is determined by deleting from the current state description all negative literals appearing in the effect list of $A$ and adding all positive literals appearing in the effect list.

As an example, the action of picking up a particular block $B$ from the floor is described in Figure 1, using the usual LISP-style notation of many planning systems. This action can be executed when the agent's hand is empty and block $B$ is clear and on the floor. After the action is executed, the agent's hand is no longer empty (it holds $B$), and $B$ is not on the floor.

Since the action of picking up a block from a location is essentially the same, regardless of the particular block and location, a whole class of such actions can be described using an *action schema* or *operator* with free variables denoting the object to be picked up and the

```
(define (action pickup-block-B-from-floor)
  :precondition (and (on floor B) (handempty) (clear B))
  :effect       (and (not (handempty)) (not (on floor B)) (holding B))))
```

Figure 1: The *Pickup-block-B-from-floor* action

```
(define (operator pickup)
  :parameters   (?x ?y)
  :precondition (and (on ?x ?y) (handempty) (clear ?x) (not (= ?x ?y)))
  :effect       (and (not (handempty)) (not (on ?x ?y)) (holding ?x))))
```

Figure 2: The *Pickup* action schema

pickup location. An action schema specification is similar to the specification of a single action except for the use of free variables. The precondition list of an action schema can contain, along with predicates (or more precisely, proposition "schemata"), equality and inequality constraints on the variables.

Figure 2 illustrates an action schema for the pickup action. It has two variables, $?x$ and $?y$, which stand for the object being picked up and the location of the object, respectively. The precondition list includes the requirements that $?x$ be on $?y$, that the hand is empty, that $?x$ is clear, and that $?x$ and $?y$ designate different objects (i.e., one cannot pickup an object from atop itself).

The STRIPS representation can be enhanced, obtaining a more expressive language that allows for a form of universal quantification in the action description (e.g., as in UCPOP Penberthy & Weld, 1992). In addition, *conditional effects* can be captured using a *when clause* consisting of an *antecedent* and a *consequent*. The semantics of the action description is similar to the original semantics except that in states $s$ that satisfy the preconditions of the action *and* the antecedent of the *when* clause, the actual effect of the action is the union of the "standard" effect specified in the effect list and the consequent of the *when* clause.

The *when* clause does not change the expressiveness of the language—each conditional action description can be expressed using separate non-conditional actions in the classic STRIPS representation to capture each *when* clause. However, it allows for a more economical and natural specification of actions. For example, in the classic STRIPS blocks world, after putting some block $B_1$ on a destination block $B_2$, block $B_2$ is no longer clear. However, after putting $B_1$ on the table, the table remains clear. Hence, a different *putdown* schema is required to describe moving a block to the table. Using a *when* clause, we can use a single schema with a conditional effect that modifies the standard effect of the action in case the destination is not the table (i.e., the *when* clause will state that when the destination is not the table, it will become unclear). In addition, conditional effects may allow us to postpone commitment during planning (e.g., we may decide to put a block down, but we don't have to commit to whether the destination is the table or not).

## 2.2 Representing Concurrent Actions in STRIPS

The introduction of concurrent interacting actions requires us to address two issues specific to the multiagent setting: *who* is performing the action, and *what other actions* are being

```
(define (operator pickup)
  :parameters    (?a1 ?x ?y)
  :precondition  (and (on ?x ?y)(handempty ?a1) (clear ?x)(not (= ?x ?y)))
  :concurrent    (not (and (pickup ?a2 ?x ?y) (not (= ?a1 ?a2))))
  :effect        (and (not (handempty ?a1)) (not (on ?x ?y)) (holding ?a1 ?x)))
```

Figure 3: The multiagent *Pickup* schema

performed at the same time. First, we deal with the identity of the performing agent by introducing an agent variable to each action schema. When the schema is instantiated, this variable is bound to a constant denoting the particular agent that is carrying out the action. Second, we must take into account the fact that for an action to have a particular effect, certain actions may or may not be performed concurrently. We capture such constraints by adding a *concurrent action list* to the existing precondition and effect lists in the specification of an action. The concurrent action list is a list of action schemata and *negated* action schemata, some of which can be partially instantiated. If an action schema $A'$ appears in the concurrent action list of an action $A$ then an instance of schema $A'$ must be performed concurrently with action $A$ in order to have the intended effect. If an action schema $A'$ appears negated in the concurrent action list of an action $A$ then no instance of schema $A'$ can be performed concurrently with action $A$ if $A$ is to have the prescribed effect.

The concurrent action list is similar to the precondition list in the following sense: when the constraints it specifies on the environment in which the action is performed are satisfied, the action will have the effects specified in the effect list. Notice that positive action schemata are implicitly existentially quantified—one instance of that schema must occur concurrently—whereas negated action schema are implicitly universally quantified—no instance of this schema should be performed concurrently.

A schema $A'$ appearing in the concurrent action list of schema $A$ can be partially instantiated or constrained: if $A'$ contains free variables appearing in the parameter list of $A$, then these variables must be instantiated as they are instantiated in $A$. In addition, constraints that restrict the possible instantiations of the schema $A$ can appear within the concurrent action list. This can be seen in the description of the multiagent setting version of the action *pickup* shown in Figure 3. The multiagent *pickup* schema has an additional parameter, $?a1$, signifying the performing agent. Its list of preconditions and effects is similar to that of the single-agent pickup schema, but it also has the concurrent action list:

$$\text{(not (and (pickup ?a2 ?x ?y) (not (= ?a1 ?a2))))}$$

The "not" prefix restricts the set of actions that can be performed concurrently with any instance of the schema $Pickup(?a1, ?x, ?y)$. In particular, we disallow concurrent execution of any instance of the schema $Pickup(?a2, ?x, ?y)$ such that $?a2$ is different from $?a1$. That is, no other agent should attempt to pickup the object $?x$ at the same time.

Using this representation, we can represent actions whose effects are modified by the concurrent execution of other actions. For example, suppose that when agent $a_1$ lifts up one side of a table all blocks on it are dumped onto the floor as long as no other agent $a_2$ lifts the other side; but if some agent $a_2$ does lift the other side of the table then the effect is simply to raise the side of the table. Clearly, we can distinguish between these two

```
(define (operator lower)
  :parameters    (?a1 ?s1)
  :precondition  (and (holding ?a1 ?s1) (raised ?s1))
  :effect        (and (not (raised ?s1))
                      (forall ?x
                        (when ((ontable ?x)
                               (not (and (lower ?a2 ?s2)(not (= ?s1 ?s2)))))
                              (and (onfloor ?x) (not (ontable ?x)))))))
```

Figure 4: The *Lower* action schema

cases using the concurrency conditions (`not (lift ?a2 ?side)`) and (`lift ?a2 ?side`). However, treating them as standard concurrency conditions essentially splits the action into two separate actions with similar effects. As in single-agent representations, we can treat such "modifiers" using a *when* clause; but now, the antecedent of the *when* clause has two parts: a list of additional preconditions and a list of additional concurrency conditions. The general form of the *when* clause is now (`when antecedent effect`), where the antecedent itself consists of two parts: (`preconditions concurrency-constraints`). The latter list has the same form as that of the concurrent-action list, and similar semantics. Thus, whenever the precondition part of the antecedent is satisfied in the current state and the concurrency condition is satisfied by the actions executed concurrently, the actual effect of the action is obtained by conjoining the standard effect with the consequent of the *when* clause.

The syntax of *when* clauses is illustrated in the table-lowering action described in Figure 4. Notice that this operator contains a universally quantified effect, that is, an effect of the form (`forall ?x (effect ?x)`). This allows us to state that the conditional effect, described by the *when* clause, applies to any object $?x$ that satisfies its precondition (e.g., to every object on the table in this case). The use of universally quantified conditional effects in finite domains is well understood (see Weld's (1994) discussion). However, to simplify our presentation, we do not treat it formally in this paper.

When we lower one side of the table, that side is no longer raised. In addition, if there is some object on the table, then lowering one side of the table will cause that object to fall, as long as the other side of the table is not being lowered at the same time. Here, we use universal quantification to describe the fact that this will happen to *any* object that is on the table. Notice that in the concurrent part of the antecedent we see a constrained schema again. It stipulates that the additional effect (i.e., the objects falling to the floor from the table) will occur if no instance of the schema $lower(?a2, ?s2)$ is executed concurrently, where $?s1$ is different than $?s2$.[3]

An action description can have no *when* clause, one *when* clause, or multiple *when* clauses. In the latter case, the preconditions of all the *when* clauses must be disjoint.[4] One might insist that the set of *when* clauses be exhaustive as well; however, we do not require this. If no *when* clause is satisfied when an action is performed, we assume that the "additional" effect is null; that is, the effect of the action is simply that given by the main effect list. When we discuss the *when* clauses of a specific action in our formal definitions below, we will generally assume the existence of an implicit *when* clause whose precondition consists of the negation of preconditions of the explicitly specified *when* clauses, and whose effect list is empty. This allows our definitions to be stated more concisely.[5]

---

3. In certain cases we might also insist that $?a1 \neq ?a2$, if agents can perform only one action at a time. But an agent with multiple effectors (to take one example) might be able to lower one or both sides concurrently. See below for more on this.

4. In the case of multiple clauses, the disjointness restriction can be relaxed if the effects are independent, much like in a Bayes net action description (Boutilier & Goldszmidt, 1996).

## 2.3 The Semantics of Concurrent Action Specifications

The semantics of individual actions is, of course, different in our multiagent setting than in the single-agent case. It is not individual actions that transform one state of the world into another state of the world. Rather it is joint actions that define state transitions. Joint actions describe the set of individual actions (some of which could be no-ops) performed by each of the agents; that is, they are $n$-tuples of individual actions.

Given a joint action $a = \langle a_1, \cdots, a_n \rangle$, we refer to the individual actions $a_i$ as the *elements* of $a$. We say that the concurrent action list of an element $a_i$ of $a$ is *satisfied* with respect to $a$ just when, for every positive schema $A$ in this list, $a$ contains some element $a_j (j \neq i)$ which is an instance of $A$, and for every negative schema $A'$ in the list, none of the elements $a_j$ $(1 \leq j \leq n)$ is an instance of $A'$. Ignoring for the moment the existence of *when* clauses, we can define the notion of joint action consistency in a straightforward manner:

**Definition** Let $a = \langle a_1, \cdots, a_n \rangle$ be a joint action where no individual action $a_i$ contains a *when* clause. We say $a$ is *consistent* if

- The precondition lists $p_i$ of each $a_i$ are jointly (logically) consistent (i.e., they do not contain a proposition and its negation).

- The effect lists $e_i$ of each $a_i$ are jointly consistent.

- The concurrent action list of each element of $a$ is satisfied w.r.t. $a$.

Given a state $s$, a consistent joint action $a = \langle a_1, \cdots, a_n \rangle$ can be executed in $s$ if the precondition lists of all elements of $a$ are satisfied in $s$. The resulting state $t$ is obtained by taking the union of the effect lists of each of the elements of $a$ and applying it to $s$, as in the single-agent case. In fact, a consistent joint action $a$ can be viewed as a single-agent action whose preconditions are the union of the preconditions of the various $a_i$ and whose effects are the union of the effects of the $a_i$.

Notice that under this semantics, a joint action is inconsistent if some individual action $a$ causes $Q$ to be true, and another $b$ causes $Q$ to be false. It is the responsibility of the axiomatizer of the planning domain to recognize such conflicts and either state the true effect when $a$ and $b$ are performed concurrently (by imposing conditional effects with concurrent action conditions) or to disallow concurrent execution (by imposing nonconcurrency conditions).[6]

---

5. We do not assume that such a clause is ever explicitly constructed for planning purposes—it is merely a conceptual device.

6. One can easily preprocess actions descriptions in order to check for consistency. If actions $a$ and $b$ are discovered to have conflicting effects, but the specification allows them to be executed concurrently, an algorithm could automatically add a nonconcurrency constraint to each action description, thus

With *when* clauses the definition of consistency is a bit more involved. Consistent joint actions without *when* clauses can be applied consistently at all possible states (if they are applicable at all). In contrast, joint actions with *when* clauses may be consistent when applied at some states, but inconsistent at others. Given a joint action $a = \langle a_1, \cdots, a_n \rangle$ and a specific state $s$, exactly one *when* clause of each action $a_i$ will be satisfied; that is, just one clause will have its preconditions and concurrency constraints satisfied.[7] Thus the joint action and the state together determine which conditional effects are selected.

**Definition** Given a joint action $a = \langle a_1, \cdots, a_n \rangle$ and state $s$, the *active when clause* $w_i$ of $a_i$ relative to $s$ and $a$ is the (unique) *when* clause that is satisfied by $s$ and $a$ (i.e., whose preconditions are satisfied by $s$ and whose concurrency constraints are satisfied by $a$).

We thus relativize the notion of consistency in this case.

**Definition** Let $a = \langle a_1, \cdots, a_n \rangle$ be a joint action (where individual actions $a_i$ may contain *when* clauses). Let $s$ be some state, let $w_i$ be the active *when* clause for $a_i$ (w.r.t. $s$, $a$), and let $w_i$ have preconditions $wp_i$, concurrency constraints $wc_i$, and effects $we_i$. We say $a$ is *consistent* at state $s$ if:

- The precondition lists $p_i$ and active *when*-preconditions $wp_i$ of each $a_i$ are mutually consistent.

- The effect lists $e_i$ and active *when*-effects $we_i$ of each $a_i$ are mutually consistent.

- The concurrent action list of each element of $a$ is satisfied w.r.t. $a$.

Note that we do not require that the concurrent action lists in the *when* clauses be satisfied, since they are "selected" by $a$. Note also that this definition reduces to the "*when*less" definition if the individual actions have no *when* clauses—an action is consistent with respect to $s$ iff it is consistent in the original sense.

Given a state $s$, a joint action $a = \langle a_1, \cdots, a_n \rangle$ (involving *when* clauses) that is consistent with respect to $s$ can be executed in $s$ if the precondition lists of all elements of $a$ are satisfied in $s$. The resulting state $t$ is obtained by taking the union of the effect lists of each of the elements of $a$, together with the effect lists of each of the active *when* clauses, and applying it to $s$.

Several interesting issues arise in the specification of actions for multiple agents. First, we assume throughout the rest of the paper that each agent can perform only one action at a time, so any possible concurrent actions must be performed by distinct agents. This allows our action descriptions to be simpler than they otherwise might. When a single agent can perform more than one action at a time, it can be captured using a group of "agents" denoting its different actuators. If these agents can only perform certain actions

---

preventing problems from arising during the planning process. This would be valid only if $a$ and $b$ could not, in fact, be (meaningfully) performed concurrently. If they can, then it is important that the domain axiomatizer specify what the true interacting effect is (e.g., maybe action $a$ dominates). We note that this automatic inconsistency detection and repair admits a certain additional degree of convenience in domain specification.

7. We assume an implicit *when* clause corresponding to the negation of explicitly stated clauses as described above.

concurrently, this can be captured by adding extra concurrency constraints. More generally, different agents may have different capabilities, and it would be useful to have the ability to explicitly specify these capabilities in the form of constraints on the types of actions that different agents can execute. One way to handle such constraints is via a preprocessing step that augments the action descriptions with additional preconditions or concurrency conditions that capture these constraints. An alternative is to alter the planning algorithm to take such constraints into account explicitly. When these are simple constraints—for instance, the fact that there are $n$ agents might imply that only $n$ actions can be executed concurrently—this can be done in a simple and efficient manner. This is the approach we take in the planning algorithm we develop in Section 3. However, when the capability constraints are complex, the former method seems better.

Another issue that must be addressed is the precise effect of a joint action, one of whose individual actions negates some precondition of a concurrently executed individual action. We make no special allowances for this, simply retaining the semantics described above. While this does not complicate the definition of joint actions, we note that some such combinations may not make sense. For example, the concurrent writing of variable $p$ to $q$ and variable $q$ to $p$ in a computer program might be seen as each action destroying the preconditions of the other; yet the net effect of the individual actions is simply a swap of values. Again, in certain circumstances, it may be acceptable to describe the actions this way, and in others this may not be the true effect of the joint action. Again, we can treat this issue in several ways: we can allow the specification of such actions and design the planner so that it excludes such combinations when forming concurrent plans unless an explicit concurrency condition is given (this means the axiomatizer need not think about such interactions); or we can allow such combinations, in general, but explicitly exclude problematic cases by adding nonconcurrency constraints.

Finally, an undesirable (though theoretically unproblematic) situation can arise if we provide "incongruous" concurrency lists. For example, we may require action $a$ to be concurrent with $b$ in order to have a particular effect, while $b$ may be required to be nonconcurrent with $a$ (this can span a set of actions with more than two elements, naturally). Hence, $a$ and $b$ cannot occur together in a consistent joint action, and we would not be able to achieve the intended effect of $a$. Although the planner will eventually "recognize" this fact, such specifications can lead to unnecessary backtracking during the planning process. Again, this is something that is easily detected by a preprocessor, and we will generally assume that concurrency lists are congruous.

## 2.4 Concurrent Plan Representation

Before moving on to discuss the planning process, we describe our representation for multi-agent plans, which is a rather straightforward extension of standard single-agent, partially ordered plan representations. A (single-agent) nonlinear plan consists of: (1) a set of action instances; (2) various strict *ordering constraints* using the relations $<$ and $>$ on the ordering of these actions; and (3) *codesignation* and *non-codesignation constraints* on the values of variables appearing in these actions, forcing them to have the same or different values, respectively (Weld, 1994; Penberthy & Weld, 1992). A plan of this sort represents its set of possible *linearizations*, the set of totally ordered plans formed from its action instances that do not violate any of the ordering, codesignation, and non-codesignation constraints.[8] We say a plan is *consistent* if it has some linearization. The set of linearizations can be seen as the "semantics" of a nonlinear plan in some sense. A (consistent) nonlinear plan *satisfies* a goal set $G$, given starting state $s$, if any linearization is guaranteed to satisfy $G$.

A *concurrent nonlinear plan* for $n$ agents (labeled $1, \ldots n$) is similar: it consists of a set of action instances (with agent arguments, though not necessarily instantiated) together with a set of arbitrary ordering constraints over the actions (i.e., $<, >, =$ and $\neq$) and the usual codesignation and non-codesignation constraints. Unlike single-agent nonlinear plans, we allow equality and inequality ordering constraints so that concurrent or nonconcurrent execution of a pair of actions can be imposed. Our semantics must allow for the concurrent execution of actions by our $n$ agents. To this end we extend the notion of a linearization:

**Definition** Let $P$ be a concurrent nonlinear plan for agents $1, \ldots n$. An $n$-*linearization* of $P$ is a sequence of joint actions $A_1, \cdots A_k$ for agents $1, \ldots n$ such that

1. each individual action instance in $P$ is a member of exactly one joint action $A_i$;

2. no individual action occurs in $A_1, \cdots A_k$ other than those in $P$, or individual *No-op* actions;

3. the codesignation and non-codesignation constraints in $P$ are respected; and

4. the ordering constraints in $P$ are respected. More precisely, for any individual action instances $a$ and $b$ in $P$, and joint actions $A_j$ and $A_k$ in which $a$ and $b$ occur, any ordering constraints between $a$ and $b$ are true of $A_j$ and $A_k$; that is, if $a\{<, >, =, \neq\}b$, then $j\{<, >, =, \neq\}k$.

In other words, the actions in $P$ are arranged in a set of joint actions such that the ordering of individual actions satisfies the constraints, and "synchronization" is ensured by no-ops. Note that if we have a set of $k$ actions (which are allowed to be executed by distinct agents) with no ordering constraints, the set of linearizations includes the "short" plan with a single joint action where all $k$ actions are executed concurrently by different agents (assuming $k \leq n$), a "strung out" plan where the $k$ actions are executed one at a time by a single agent, with all others doing nothing (or where different agents take turns doing the individual actions), "longer" plans stretched out even further by joint no-ops, or anything in between.

**Example** Suppose our planner outputs the following plan for a group of three agents: the set of actions is

$$\{a(1), b(2), c(2), d(3), e(1), f(2)\}$$

with the ordering constraints

$$\{e(1) = b(2), c(2) \neq d(3), a(1) < e(1), d(3) < f(2)\}$$

Here, the numerical arguments denote the agent performing the action. Joint actions involve one action for each of the three agents. A simple 3-linearization of this plan—depicted as the first linearization in Figure 5, and using $N$ to denote no-ops for the

---

8. Concurrent execution has also been considered in this context for non-interacting actions; see Knoblock's discussion of this issue (Knoblock, 1994).

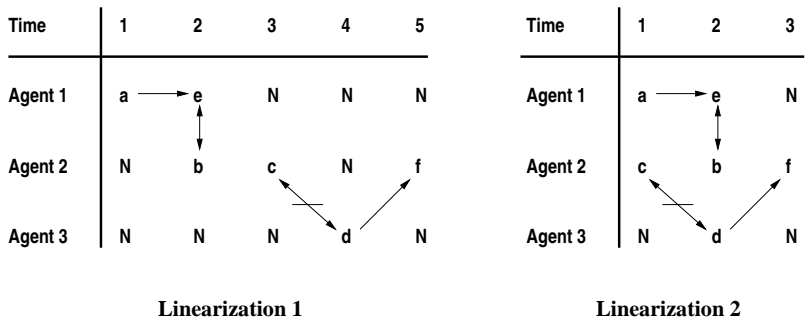

Figure 5: Two possible linearizations of a partially ordered multiagent plan

corresponding agents—is:

$$\langle a(1), N(2), N(3) \rangle, \langle e(1), b(2), N(3) \rangle, \langle N(1), c(2), N(3) \rangle, \langle N(1), N(2), d(3) \rangle, \langle N(1), f(2), N(3) \rangle$$

We can insert additional tuples of the form $\langle N(1), N(2), N(3) \rangle$ in any location we wish. Another possible 3-linearization (the second in Figure 5) is:

$$\langle a(1), c(2), N(3) \rangle, \langle e(1), b(2), d(3) \rangle, \langle N(1), f(2), N(3) \rangle$$

In fact, this is the shortest 3-linearization of the plan.

The definition of $n$-linearization requires that no agent perform more than one action at a time. This conforms with the assumption we made in the last section, though the definition could quite easily be relaxed to allow this. Because of no-ops, our $n$-linearizations do not correspond to shortest plans, either in the concurrently on nonconcurrently executed senses of the term. However, it is a relatively easy matter to "sweep through" a concurrent nonlinear plan and construct some *shortest $n$-linearization*, one with the fewest joint actions, or taking the least amount of "time." Though we do not have an explicit notion of time, the sequence of joint actions in an $n$-linearization implicitly determines a time line along which each agent must execute its individual actions. The fact that concurrency and nonconcurrency constraints are enforced in the linearizations ensure that the plan is coordinated and synchronized. We note that in order to *execute* such a plan in a coordinated fashion the agents will need some synchronization mechanism. This issue is not dealt with in this paper.

## 3. Planning with Concurrent Actions

In Figure 6, we present the POMP algorithm, a version of Weld's POP algorithm (Weld, 1994) modified to handle concurrent actions. To keep the discussion simple, we begin by describing POMP without considering conditional action effects. Below we describe the simple modifications required to add conditionals (i.e., to build the analog of CPOP). Though we do not discuss universal quantification in this paper, our algorithm could easily be extended to handle universally quantified effects in much the same way as Penberthy and Weld's (1992) full UCPOP algorithm.

<u>POMP($\langle A, O, L, NC, B \rangle$, *agenda*)</u>

**Termination:** If agenda is empty, return $\langle A, O, L, NC, B \rangle$.

**Goal Selection:** Let $\langle Q, A_{need} \rangle$ be a pair on the *agenda*. ($A_{need}$ is an action and $Q$ is a conjunct from its precondition list.)

**Action Selection:** Let $A_{add} = Choose$ an action one of whose effects unifies with $Q$ subject to the constraints in $B$. (This may be a newly instantiated action from $\Lambda$ or an action that is already in $A$ and can be ordered consistently prior to $A_{need}$). If no such action exists, then return failure. Let $L' = L \cup \{A_{add} \overset{Q}{\to} A_{need}\}$. Form $B'$ by adding to $B$ any codesignation constraints that are needed in order to force $A_{add}$ to have the desired effect. Let $O' = O \cup \{A_{add} < A_{need}\}$. If $A_{add}$ is newly instantiated, then $A' = A \cup \{A_{add}\}$ and $O' = O' \cup \{A_0 < A_{add} < A_\infty\}$ (otherwise, let $A' = A$).

**Concurrent Action Selection:** If $A_{add}$ is newly instantiated then apply the following steps to all positive actions $\alpha_{conc}$ in its concurrent list: Let $A_{conc} = Choose$ a newly instantiated action from $\Lambda$ or an action that is already in $A$ and can be ordered consistently concurrently with $A_{add}$. Make sure that there is a free agent that can perform this action concurrently with $A_{add}$ and any other concurrently scheduled actions. If no such action exists then return failure. Let $O' = O \cup \{A_{conc} = A_{need}\}$. If $A_{conc}$ is newly instantiated, then $A' = A \cup \{A_{add}\}$ and $O' = O' \cup \{A_0 < A_{conc} < A_\infty\}$ (otherwise, let $A' = A$). If $a_{add}$ is the agent variable in $A_{add}$ and $a_{conc}$ is the agent variable in $A_{conc}$, then add $a_{add} \neq a_{conc}$ to $B'$, as well as all similar non-codesignation constraints for actions $A$ such that $A = A_{add} \in O$.

Re-apply this step to $A_{conc}$, if needed.

For every negative action $A_{\neg conc}$ in $A_{add}$ concurrent list let $NC' = NC \cup \{A_{\neg conc} \neq A_{add}\}$. Add to $B'$ any codesignation constraints associated with $A_{\neg conc}$.

**Updating of Goal State:** Let $agenda' = agenda - \{\langle Q, A_{need} \rangle\}$.

If $A_{add}$ is newly instantiated, then add $\{\langle Q_j, A_{add} \rangle\}$ to $agenda'$ for every $Q_j$ that is a logical precondition of $A_{add}$. Add the other preconditions to $B'$. If additional concurrent actions were added, add their preconditions as well.

**Causal Link Protection:** For every action $A_t$ that might threaten a causal link $A_p \overset{R}{\to} A_c$ perform one of

  (a) Demotion: Add $A_t < A_p$ to $O'$.

  (b) Weak Promotion: Add $A_t \geq A_c$ to $O'$. If no agent can perform $A_t$ concurrently with $A_c$, add $A_t > A_c$, instead.

If neither constraint is consistent, then return failure.

**Nonconcurrency Enforcement** For every action $A_t$ that threatens a nonconcurrency constraint $\mathsf{A} \neq A$ (i.e., $A_t$ is an instance of the schema $\mathsf{A}$ that does not violate any constraint in $B'$) add a consistent constraint, either

  (a) Demotion: Add $A_t < A$ to $O'$.

  (b) Promotion: Add $A_t > A$ to $O'$.

If neither constraint is consistent, then return failure.

**Recursive Invocation:** POMP($\langle A', O', L', NC', B' \rangle$, *agenda'*)

Figure 6: The **P**artially **O**rdered **M**ultiagent **P**lanning algorithm

We assume the existence of a function $MGU(Q, R, B)$ which returns the most general unifier of the literals $Q$ and $R$ with respect to the codesignation constraints in $B$. This is used wherever unification of action schemata is required (see the Action Selection step in Figure 6 and our discussion of NC-threats below). The algorithm has a number of input variables: the set $A$ contains all action instances inserted into the plan so far; the set $O$ contains ordering constraints on elements of $A$; the set $L$ contains causal links; the set $NC$ contains nonconcurrency constraints; and the set $B$ contains the current codesignation constraints. The set $NC$ does not appear in the POP algorithm and contains elements of the form $\mathsf{A} \neq a$, where $\mathsf{A} \in \Lambda$ is an action schema and $a$ is an action instance from $A$. Intuitively, a nonconcurrency constraint of this form requires that no action instance $a'$ that matches the schema $\mathsf{A}$ subject to the (non) codesignation constraints should appear concurrently with $a$ in the plan.

The *agenda* is a set of pairs of the form $\langle Q, A \rangle$, listing preconditions $Q$ that have not been achieved yet and the actions $A$ that require them. Initially, the sets $L$, $NC$, and $B$ are empty, while $A$ contains the two fictitious actions $A_0$ and $A_\infty$, where $A_0$ has the initial state propositions as its effects and $A_\infty$ has the goal state conditions as its preconditions. The agenda contains all pairs $\langle Q, A_\infty \rangle$ such that $Q$ is one of the conjuncts in the description of the goal state. This specification of the initial agenda is identical to that used in POP (Weld, 1994). Finally, we note that the *choose* operator, which appears in the **Action Selection** and **Concurrent Action Selection** steps, denotes nondeterministic choice. Again, this device is just that used in POP to make algorithm specification independent of the search strategy actually used for planning. Intuitively, a complete planner will require one to search over nondeterministic choices, backtracking over those that lead to failure.

Many of the structures and algorithmic steps of POMP correspond exactly to those used in POP. Rather than describe these in detail, we focus our discussion on the elements of POMP that differ from POP. Apart from the additional data structure $NC$ mentioned above, one key difference is the additional **Concurrent Action Selection** step in POMP, which takes care of the concurrency requirements of each newly instantiated action.

One final key distinction is the notion of a *threat* used in POMP, which is more general than that used by POP. Much like POP, given a plan $\langle A, O, L, NC \rangle$, we say that $A_t$ *threatens* the causal link $A_p \xrightarrow{Q} A_c$ when $O \cup \{A_p \leq A_t < A_c\}$ is consistent, and $A_t$ has $\neg Q$ as an effect. Threats are handled using *demotion* (much like in POP), or *weak promotion*. The latter differs from the standard promotion technique used in POP: it allows $A_t$ to be ordered *concurrently* with $A_c$, not just after $A_c$.[9]

Apart from handling conventional threats in a different manner, we have another form of threat in concurrent plans, namely, *NC-threats*. We say that action instance $A_t$ *threatens* the nonconcurrency constraint $\mathsf{A} \neq A_c$ if $O \cup \{A_t = A_c\}$ is consistent and $A_t$ is an instantiation of $\mathsf{A}$ that does not violate any of the codesignation constraints. Demotion and promotion can be used to handle NC-threats, just as they do more conventional threats. Notice that although the set $NC$ contains negative (inequality) constraints, they will ultimately be grounded in the set of positive constraints in $O$. Following the approach suggested by Weld (1994), we do not consider an action to be a threat if some of its variables can be consistently instantiated in a manner that would remove the threat.

---

9. If we wish to exclude actions that negate some precondition of another concurrent action (see discussion in Section 2), we must use $O \cup \{A_p \leq A_t \leq A_c\}$ in the definition of threat, and we must change weak promotion to standard promotion.

The POMP algorithm must check for the consistency of ordering constraints in several places: in **Action Selection** where an action chosen to achieve an effect must be consistently ordered before the consumer of that effect; in **Concurrent Action Selection** where each concurrency requirement added to the plan must be tested for consistency; and in **Nonconcurrency Enforcement** where demotion or promotion is used to ensure that no nonconcurrency requirements are violated. The consistency testing of a set of ordering constraints is very similar to that employed in POP (see Weld (1994) for a nice discussion), with one key difference: the existence of equality ($=$) and inequality ($\neq$) ordering constraints as opposed to simple strict inequalities (i.e., $<$ and $>$). However, with minor modifications, standard consistency-checking algorithms for strict ordering constraints can be used. Equality can be dealt with by simply "merging" actions that must occur concurrently (i.e., treating them as a single action for the purposes of consistency testing). Inequalities are easily handled by assuming all actions occur at different points whenever possible. Non-strict inequalities (i.e., $\leq$ and $\geq$) do not arise directly in our algorithm (though these two can be easily dealt with). We refer to Ghallab and Alaoui (1989) for further details on processing such constraints.

The POMP algorithm as described can easily be modified to handle conditional effects, just as the POP algorithm can be extended to CPOP. The main fact to note is that in the action selection phase, we can use an action whose conditional effects achieve the chosen subgoal. In that case, we do not just add the preconditions of the selected action to the agenda, but also the antecedent of the particular conditional effect (this to ensure that the action has this particular effect). We handle the additional concurrency conditions in the antecedent much like the regular concurrency conditions. As in the CPOP algorithm, we must consider the possibility that a particular conditional effect of an added action threatens an established causal link. In this case, we can, aside from using the existing threat resolution techniques, consider a form of *confrontation*, where we add the negation of the conditional effect's antecedent to the agenda. Again, we have several ways to do this: we could add the negation of some literal in the antecedent's condition to the agenda; but we can also *add a concurrent action* to negate a negative concurrency condition in the antecedent, or *post a nonconcurrency constraint* to offset a positive concurrency constraint in the antecedent. The details of such steps are straightforward and look similar to those involved in the unconditional algorithm.

## 4. An Example of the POMP Algorithm

In this section, we formalize the example alluded to in the introduction and describe the construction of a concurrent plan for this problem using the POMP algorithm.

In the initial state, two agents, *Agent1* and *Agent2*, are located in *Room1*, together with a table and a set of blocks scattered around the room. Their goal is to ensure that all of the blocks are in *Room2* and the table is on the floor. In order to simplify this example, we assume there is only one block $B$, we omit certain natural operators, and we simplify action descriptions. In order to *compactly* represent the multiple block version of this, we would require the introduction of universal quantification. As shown by Weld (1994), this can be done with little difficulty. Intuitively, the agents should gather the blocks in the room (in this case only one), put them on the table, carry the table to the other room, dump the blocks from the table, and then put the table down. While this is not the best plan for a single block, it illustrates how such a plan would be constructed for multiple blocks (in which case this strategy is better than that of agents making multiple trips carrying individual blocks). We use the following actions:

- *Pickup(a, b)*: agent *a* picks up a block *b*
- *PutDown(a, b)*: agent *a* puts block *b* on the table
- *ToTable(a, s)*: agent *a* moves to side *s* (left, right) of the table
- *MoveTable(a, r)*: agent *a* moves to room *r* with the table
- *Lift(a, s)*: agent *a* lifts side *s* of the table
- *Lower(a, s)*: agent *a* lowers side *s* of the table

The *a* variables are of type *agent*, *b* variables are of type *block*, *r* variables are of type *room*, and *s* variables are of type *table-side*. (We omit other natural actions since they won't be used in the plan of interest.)

The domain is described using the following predicates:

- *OnTable(b)*: block *b* is on the table
- *OnFloor(b)*: block *b* is on the floor
- *AtSide(a, s)*: agent *a* is at side *s* (left, right) of the table
- *Up(s)*: side *s* of the table is raised
- *Down(s)*: side *s* of the table is on the floor
- *InRoom(x, r)*: object *x* (agent, block, table) is in room *r*
- *HandEmpty(a)*: the hand of agent *a* is empty
- *Holding(a, x)*: agent *a* is holding *x* (block, side of table)

The operator descriptions are defined in Figure 7.

The initial state of our planning problem is:

$$\{InRoom(B, Room1),\ OnFloor(B),\ InRoom(Agent1, Room1),\ InRoom(Agent2, Room1),$$
$$InRoom(Table, Room1),\ Down(LeftSide),\ Down(RightSide)\}$$

The goal propositions are:

$$\{InRoom(B, Room2),\ OnFloor(B),\ Down(LeftSide),\ Down(RightSide)\}$$

We now consider how a concurrent nonlinear plan for this multiagent planning problem might be generated by POMP.

```
(define (operator pickup)
  :parameters    (?a1 ?x)
  :precondition  (and (inroom ?a1 ?r1) (inroom ?x ?r1)
                      (handempty ?a1) (onfloor ?x))
  :concurrent    (and (not (pickup ?a2 ?x)) (not (= ?a1 ?a2)))
  :effect        (and (not (handempty ?a1)) (not (onfloor ?x)) (holding ?a1 ?x)))

(define (operator putdown)
  :parameters    (?a1 ?x)
  :precondition   (and (inroom ?a1 ?r1) (inroom ?x ?r1) (inroom Table ?r1)
                      (holding ?a1 ?x))
  :concurrent    (not (lift ?a2 ?s1))
  :effect        (and (not (holding ?a1 ?x)) (ontable ?x) (handempty ?a1)))

(define (operator totable)
  :parameters    (?a1 ?s1)
  :precondition  (and (inroom ?a1 ?r1) (inroom Table ?r1) (not (atside ?a2 ?s1)))
  :concurrent    (and (not (totable ?a2 ?s1)) (not (= ?a1 ?a2)))
  :effect        (atside ?a1 ?s1))

(define (operator movetable)
  :parameters    (?a1 ?r1)
  :precondition  (holding ?a1 Table)
  :concurrent    (and (movetable ?a2 ?r1) (not (= ?a1 ?a2)))
  :effect        (and (inroom ?r1 Table) (inroom ?r1 ?a1)
                      (when ((ontable ?x) ()) (inroom ?r1 ?x))))

(define (operator lower)
  :parameters    (?a1 ?s1)
  :precondition  (and (holding ?a1 ?s1) (up ?s1))
  :concurrent    (and (not (lift ?a2 ?s2)) (not (= ?a1 ?a2)) (not (= ?s1 ?s2)))
  :effect        (and (not (up ?s1))(down ?s1) (not (holding ?a1 ?s1))
                      (when ((and (ontable ?x) (up ?s2) (not (= ?s1 ?s2)))
                            (and (not (lower ?a2 ?s2)) (not (= ?a2 ?a1))))
                            (and (onfloor ?x) (not (ontable x)))))))

(define (operator lift)
  :parameters    (?a1 ?s1)
  :precondition  (and (atside ?s1 ?a1) (down ?s1) (down ?s2) (not (= ?s1 ?s2)))
  :concurrent    (and (not (lower ?a2 ?s2)) (not (= ?a1 ?a2)) (not (= ?s1 ?s2)))
  :effect        (and (not (down ?s1)) (up ?s1) (holding ?a1 ?s1)
                      (when ((and (ontable ?x) (down ?s2) (not (= ?s1 ?s2)))
                            (and (not (lift ?a2 ?s2))))
                            (and (onfloor ?x) (not (ontable x)))))))
```

Figure 7: The table movers domain

Suppose that $InRoom(B, Room2)$ is the first goal selected. This can be achieved by performing $A_1 = MoveTable(a1, Room2)$ via its conditional effect (note that $a1$ is an agent variable, so there is no commitment to which agent performs this action).[10] We must add both $Holding(a1, Table)$ and $OnTable(B)$ to the agenda and insert the appropriate causal links. In addition, the concurrent list forces us to add the action $A_2 = MoveTable(a2, Room2)$ to the plan together with the non-codesignation constraint $a1 \neq a2$. The ordering constraint $A_1 = A_2$ is added as well. When we add $A_2$, we must add its precondition $Holding(a2, Table)$ to the agenda as well. The structure of the partially constructed plan might be viewed as follows:[11]

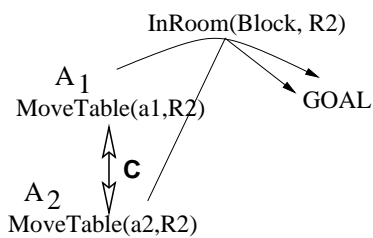

Next, we choose the subgoal $OnTable(B)$ from the agenda (which we just added). We add the action $A_3 = PutDown(a3, B)$ to the plan with the appropriate ordering constraint $A_3 < A_1$; its preconditions are added to the agenda and a causal link is added to $L$. In addition, we must add to $NC$ the nonconcurrency constraint $not(Lift(a, s))$: no agent can lift any side of the table while the block is being placed on it if the desired effect is to be achieved.

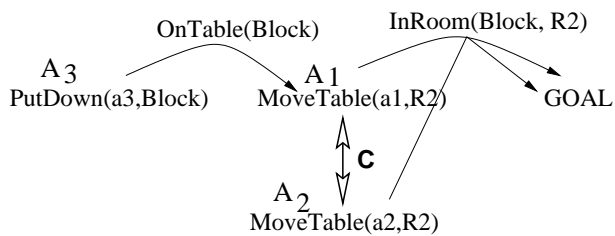

---

10. We do not pursue the notion of heuristics for action selection here; but we do note that this action is a plausible candidate for selection in the multi-block setting. If the goal list asserts that a number of blocks should be in the second room, the single action of moving the table will achieve all of these under the appropriate conditions (i.e., all the blocks are on the table). If action selection favors (conditional) actions that achieve more goals or subgoals, this action will be considered before the actions needed for "one by one" transport of the blocks by the individual agents. So this choice is not as silly as it might seem in the single-block setting.

11. In the plan diagrams that follow, we indicate actions as $A_i$ with the name of the action below it. Variables are indicated by lower-case names (we do not indicate co-designation constraints in the diagrams). An arrow from one action to another denotes a causal link (from producer to consumer), labeled by the proposition being produced. Large arrows labeled with a **C** (resp. **NC**) denote concurrency (resp. nonconcurrency) constraints between actions. We use left-to-right ordering to denote the temporal ordering of actions, if such constraints exist.

Now we choose the subgoal *Holding*(*a*1, *Table*). This can be achieved using $A_4 = Lift(a1, s1)$, with the ordering constraint $A_4 < A_1$. All the preconditions are added to the agenda, but no concurrency conditions are added (yet!) for this action, since we do not yet need to invoke the conditional effects of that action induced by simultaneous lifting of the other side of the table:

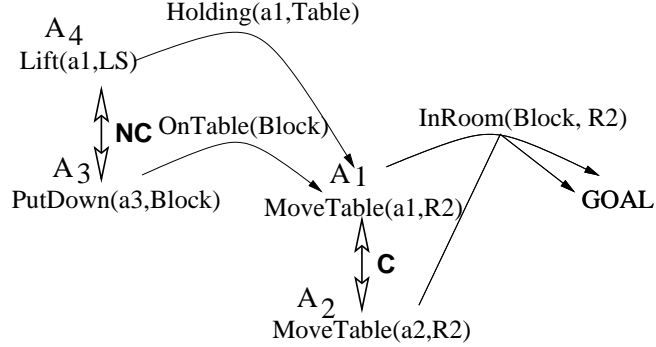

We now note that the conditional effect of $A_4$ poses a threat to the causal link $A_3 \overset{ontable}{\to} A_1$; this is because lifting a *single* side of the table will dump the block from the table. In addition, the nonconcurrency constraint associated with $A_3$, that no lifting be performed concurrently with $A_3$, is threatened by $A_4$ (an NC-threat), as indicated in the plan diagram above. The confrontation strategy is used to handle the first threat, and the action $A_5 = Lift(a4, s2)$ scheduled concurrently with $A_4$. The constraints $s1 \neq s2$ and $a4 \neq a1$ are also imposed. This ensures that the undesirable effect will not occur. We resolve the NC threat by ordering $A_3$ before $A_4$.[12] The resulting partially completed plan is now free of threats:

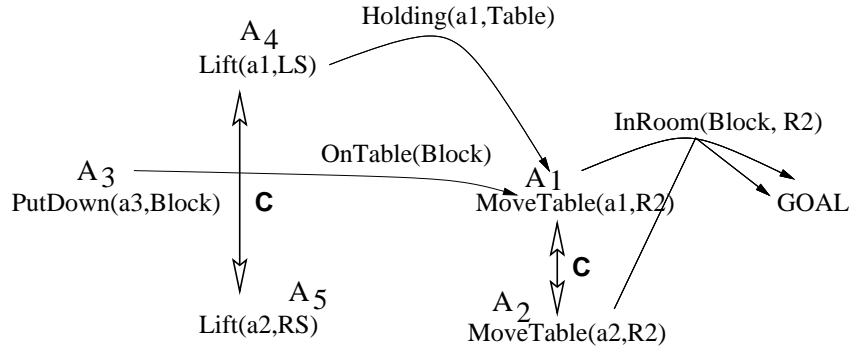

Next, we choose the subgoal *Down*(*LeftSide*). This is achieved using the action $A_6 = Lower(a1, LeftSide)$ and its preconditions are added to the agenda. In a completely similar way, $A_7 = Lower(a2, RightSide)$ is added to achieve *Down*(*RightSide*) (again, we anticipate the unification of these agent variables).

---

12. In anticipation of a subsequent step, we use variable $a2$ in the plan diagram instead of $a4$, since they will soon be unified. To keep things concrete, we have also replaced $s1$ and $s2$ with particular sides of the table, *LeftSide* and *RightSide*, to make the discussion a bit less convoluted.

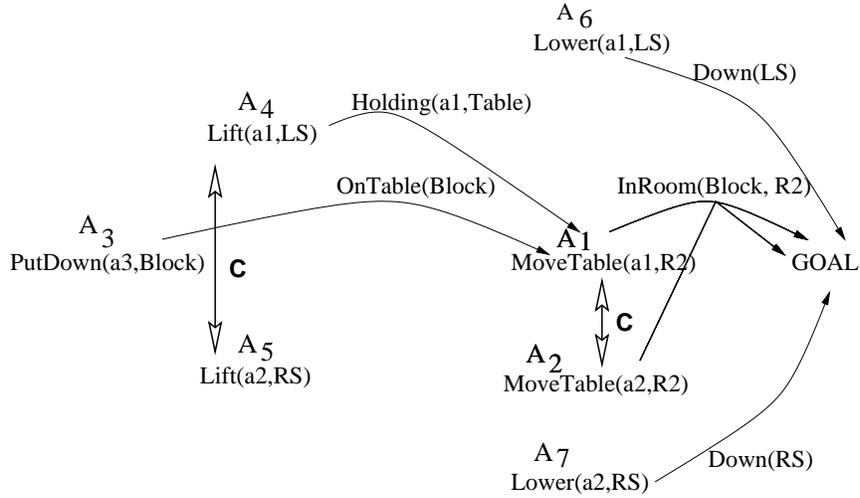

We now choose to work on the preconditions of $A_6$ and $A_7$. Both of the preconditions, *Up(s)* and *Holding(a, s)*, are effects of *Lift*, so we use $A_4$ and $A_5$ as their producers. At this stage, both $A_6$ and $A_7$ are constrained to follow $A_4$ and $A_5$, but there are no constraints on the relative ordering of $A_6$ and $A_7$ themselves. We also see that both $A_6$ and $A_7$ "potentially" threaten the causal link $A_3 \overset{ontable}{\rightarrow} A_1$; that is, they each have a conditional effect that would cause the block to fall from the table. There are several ways to resolve these two threats, including confrontation. We choose strict promotion, and order both $A_6$ and $A_7$ to occur after $A_1$ and $A_2$.

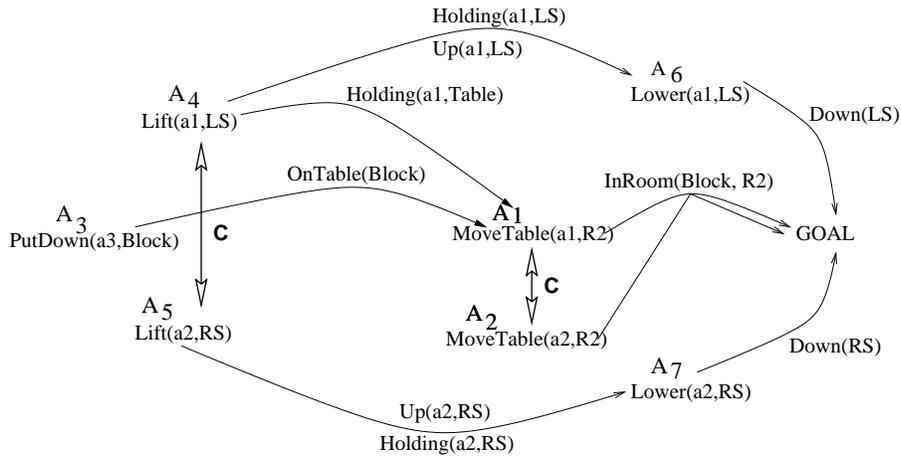

Now, we choose the subgoal *OnFloor(B)*, which is a conditional effect of the *Lower* action. We choose to accomplish it using an existing action, $A_6$. In order to obtain the desired effect, we ensure the antecedent of the when clause for this effect holds: this involves adding the conditions of the antecedent (*OnTable(B)* and *Up(LeftSide)*) to the agenda, and imposing the nonconcurrency constraint of the antecedent, namely, that no concurrent *Lower* action can take place. This constraint is threatened by the action $A_7$, so we order $A_6$ before $A_7$ by posting the constraint $A_6 < A_7$.[13] The conditions of the antecedent, $OnTable(B)$ and $Up(LeftSide)$, can use $A_3$ and $A_5$ as the producers, respectively.

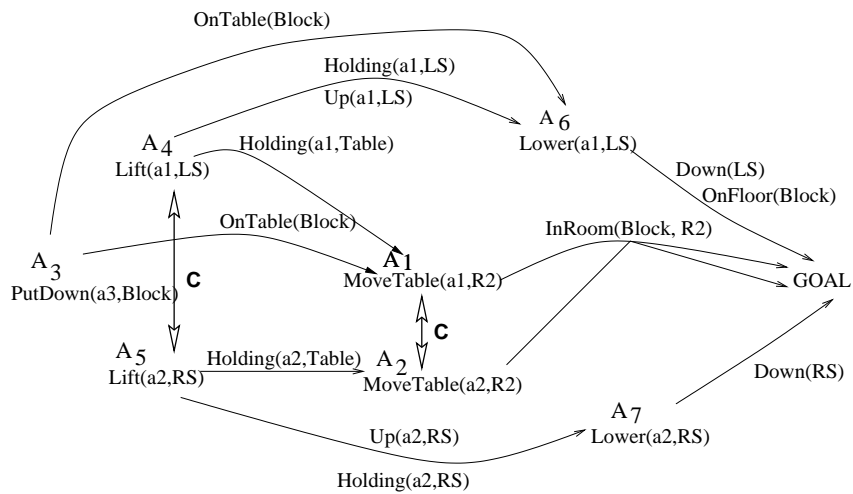

The only unsolved subgoal is the precondition of the initial $PutDown(a3, B)$ action (others, such as $Down(LeftSide)$ for the $Lift$ action, are produced by the initial state). We don't illustrate it, but it is a simple matter to introduce the $Pickup(a3, B)$ action before $PutDown(a3, B)$.

We now have the following plan: first, the block is picked up and put on the table by some agent $a3$ (either of $Agent1$ or $Agent2$ can do this). This is followed by two concurrent lift actions and two concurrent move actions which get the table to the other room with the block on top. Next, we have a single lower action, which makes the block fall off, followed by another lower action which ensures that both sides of the table are on the floor. We note that the plan does not care which of the agents (the one who lifts the $LeftSide$ or the $RightSide$) initially puts the block on the table.[14]

## 5. Soundness and Completeness of the POMP Algorithm

We say that a planning algorithm is *sound* if it generates only plans that are guaranteed to achieve the goals posed to it; a *complete* algorithm is guaranteed to generate a plan if a successful plan exists.[15] In the case of concurrent nonlinear plans, we will say that an algorithm is sound if each $n$-linearization of the plan produced for a given problem will reach a goal state, and an algorithm is complete if it successfully generates a concurrent nonlinear plan whenever there is a sequence of joint actions (i.e., an $n$-linearization of some concurrent plan) that achieves the goal from the initial state. We now show that the POMP algorithm is both sound and complete.

---

13. The other ordering $A_7 < A_6$ could have been used to resolve this threat; but it would cause an "unresolvable" threat to the conditions of the antecedent, which require that the other side remain up. It is, of course, only "unresolvable" in the sense that it would require the agents to pick up the block, etc., essentially introducing a cycle in the plan.

14. Further examples of MAP problems, the plans produced by POMP, and code implementing the POMP algorithm can be obtained at `http://www.cs.bgu.ac.il/∼ishayl/project/`.

15. For formal definitions of these concepts, we refer the reader to (Penberthy & Weld, 1992).

The soundness proof is straightforward. Suppose that the generated plan is not sound. Thus, some $n$-linearization of the plan does not achieve the goal or some required subgoal (i.e., a precondition of one of the plan's actions). Because of the agenda mechanism, it is clear that for each needed goal or precondition there exists an action in the plan that achieves that subgoal (goal or precondition). Moreover, there is an explicit causal link in the plan for that particular subgoal as well as an ordering constraint requiring that the producing action to appear prior to the consuming action (or the goal). Any $n$-linearization of a plan is another plan obtained from the original plan by adding new, *consistent*, strict (i.e. $<, >$) ordering constraints. Recall that the original plan's ordering constraints must have been consistent, otherwise it would not constitute a solution, and that there were no threats. Clearly, by adding new *strict* ordering constraints we cannot cause any new threats to causal links or violate a nonconcurrency constraint. Hence, the resulting $n$-linearization respects all causal links of the original plan and all ordering constraints of the original plan.

To complete the proof, we must be convinced that POMP actually considers all possible, relevant interactions between actions. Consider some effect $P$ of an action $a$ needed by some action $b$ which is ordered after $a$. Given the semantics of actions, there are only two reasons why $P$ will not hold prior to the execution of $b$: (1) some action $c$ between $a$ and $b$ (possibly concurrent with $a$) has an effect $\neg P$; or (2) $a$ did not actually have $P$ as an effect. Case (1) contradicts the fact that there are no threats (in our extended sense, covering the possibility of $c$ occurring concurrently with $a$) in the context of this plan. Case (2) implies that either $P$ is an effect of $a$ subject to some concurrency or nonconcurrency condition that is violated in this $n$-linearization. Any such problem would have been taken care of by the **Action Selection** or **Nonconcurrency Enforcement** steps (and by the ordering constraints). Thus it should be clear that any $n$-linearization of a plan produced by POMP does in fact achieve all its goals; that is, POMP is sound.

The completeness proof rests on three key elements:

1. A reduction from multiagent planning problems to single agent planning problems.

2. The fact that POMP can solve a multiagent planning problem iff POP can solve the single agent planning problem obtained via this reduction.

3. The fact that POP is sound and complete (Penberthy & Weld, 1992).

First, we show how given a multiagent planning problem, a similar single agent planning problem can be obtained. We shall refer to the generated problem as the *equivalent single agent planning problem* (or the *ESA problem*). This reduction has the property that a plan for the multiagent planning problem exists if and only if a plan for the ESA problem exists. In the introduction, we discussed such a reduction via the use of joint actions. Here, we will use a similar idea, but with a little more care so that both POMP and POP will perform similar steps in the solution of the original problem and the ESA problem, respectively. Combining these results with the fact that POP is sound and complete, we can deduce that POMP is sound and complete as well.

In the discussion below, we ignore conditional effects to avoid undue and, for the most part, uninteresting complications. The extension of the arguments to deal with conditional effects is straightforward. We first recall the following facts relevant to our argument:

(a) POP and POMP are nondeterministic planning algorithms and, although there are various ways of making them deterministic, this issue is orthogonal to the proof. Thus, in showing the correspondence between POP and POMP alluded to in point (2) above, we can utilize the flexibility awarded to us by each planner's use of nondeterministic choice. In particular, it is sufficient to show that for a given solution path for one planner, a similar solution path exists for the other.

(b) The choice of the next agenda element to work on is immaterial for both POP and POMP—it can affect the running time (e.g., by causing backtracking) but not the existence of a solution. Hence, we are flexible in ordering the subgoals achieved, as long as we respect causality (i.e., we cannot achieve a goal that is derived from a precondition of an action that was not introduced yet).

(c) By introducing additional ordering constraints consistent with current constraints in a valid plan, we obtain a valid plan for the given problem.

(d) The precise order in which actions and ordering constraints are inserted does not affect the validity of the solution. In fact, as is well known in the planning community, one can postpone the threat resolution step without affecting the soundness or completeness of the algorithm, as long as all threats are eventually resolved.

Our proof will proceed in two stages. In the first stage, we will limit ourselves to a restricted set of planning problems for which we can show the connections with POP in a straightforward fashion. We then relax this restriction to show the correspondence between the two planners in the general case.

Recall that in Section 2.3 we suggested a possible restriction on the set of actions one is allowed to execute concurrently, namely, that no two actions $a$ and $b$ are permitted to occur concurrently if one's effects negate any of the other's preconditions. We remarked that this *concurrent, non-clobbering condition*, if not enforced in the action specification itself, is easily enforced by the POMP algorithm if we modify the definition of a threat and use promotions instead of weak promotions to resolve threats. Let us restrict attention, for the time being, to domains respecting this condition.

We first note the following fact. Let $M$ be some POMP plan, and consider some $n$-linearization of $M$ in which $a_1$ and $a_2$ occur concurrently, but where $M$ is such that no future actions require the effects produced by the concurrent execution of these actions. That is, actions $a_1$ and $a_2$ are not *forced* to occur concurrently by plan $M$. In this case, any similar $n$-linearization in which $a_1$ is ordered before $a_2$, or vice versa, and no other ordering constraints are violated (some such linearization must exist) will also achieve the goal. The only case in which this might not happen is when one of $a_1$ or $a_2$ clobbers the other's preconditions; but this has been explicitly disallowed in our restricted setting (by the imposition of a nonconcurrency constraint or "precondition").

Now consider the ESA problem, where the actions available to the agent are as follows: for each *individual action a that has no concurrency constraints* in the multiagent problem, we create an action corresponding to the joint action where $a$ is performed by its "owning" agent, and no-ops are executed by every other agent; and for each *individual action a that has concurrency conditions*, requiring that actions $b_1, \cdots b_k$ be executed concurrently, we create an action corresponding to the joint action where $a$ and each of the $b_i$ are performed, but no other actions apart from no-ops are performed.[16] We note that nonconcurrency constraints are ignored in the ESA problem definition.

Clearly, if a joint action sequence exists for a given problem, there also exists a concurrent nonlinear plan for that problem. In addition, by the argument above involving the assumption that no concurrent action clobber another's precondition, it is also easy to see that, if a concurrent nonlinear plan can be found for a problem, there also exists a concurrent nonlinear plan in which the *only* concurrency constraints involve actions whose specification requires the concurrent execution of another action (or set of actions) in order to obtain a particular effect. This implies that, should a problem be solvable, it is solvable by a sequence of joint actions of the type constructed above, using only single-agent individual actions together with a set of no-ops, or at most involving minimal sets of interacting actions. In other words, a concurrent nonlinear plan exists for a given problem iff a plan for the ESA problem exists. We note that the structure of any solution for the ESA problem (or any linearization of a nonlinear single-agent plan for the ESA problem) is very specific: actions occur concurrently only if they are forced to. In other words, solutions to the ESA problem are *strung out plans*, in which agents "take turns" performing their actions.

Next, we want to show that (in our restricted setting) POMP's solution path for a given planning problem and POP's solution path for its ESA problem resemble each other. This becomes apparent once we combine POMP's action selection and concurrent action selection steps. We obtain a step that is equivalent to the action selection step of POP for the ESA problem (i.e., whenever POMP chooses an action which requires another concurrent action, the required concurrent action is immediately inserted as well; this is equivalent to inserting the proper ESA action). In fact, now POP and POMP look almost identical, except for POMP's **Nonconcurrency Enforcement** step. However, because of the fashion in which the ESA problem was defined, all nonconcurrency constraints are automatically "imposed" in the plan produced by POP since they refer to different joint actions. Any linearization of these joint actions enforces the nonconcurrency of all joint actions. Therefore, the only (single-agent) actions that can occur together in POP's solution to the ESA problem are those that have to occur together and on which there is no nonconcurrency constraint. (In fact, on these actions there is an explicit concurrency constraint.)[17]

The above argument demonstrates that POP and POMP generate "identical" sets of plans, except for two small differences. First, POMP's semantics allows for concurrent execution of certain actions, even though they need not be executed concurrently in order to solve the problem, while POP (for the ESA problem) cannot generate plans that admit this. However, this difference cannot affect the completeness argument (since it means that POMP is more flexible than POP).[18] Second, POMP commits to a particular ordering of actions for which there is a nonconcurrency constraint, while POP will not make such a commitment if both orderings are consistent. However, if both are consistent (and remain unordered in the final plan for the ESA problem) then the choice POMP makes cannot impact the solution (and POMP can produce either alternative if the ordering does matter). Now, using the fact that POP is sound and complete, the virtual equivalence of POMP and POP steps, and our facts about strung out plans and the ESA problem, we see that POMP is sound and complete for the special case where concurrent actions do not destroy each other's preconditions.

---

16. It is important to note that a single action schema gives rise to $n$ individual actions, one for each agent (e.g., *Lift*($Agent1, s$) and *Lift*($Agent2, s$) are distinct actions, and separate joint actions for these will be created). Similarly, when the concurrency conditions involve action schemata, any permitted combination of agent instantiations will give rise to a distinct joint action.

17. This assumes that concurrency lists are *congruous*, as described in Section 2; but if, not, a simple redefinition of the ESA problem can be given so that no "incongruous" concurrent actions are admitted.

Finally, we wish to remove the restrictions placed on concurrent actions, and admit problems where a concurrent action can clobber the precondition of another. We note that problems of this type exist that cannot be solved by a *strung out* plan in the sense defined above. For instance, consider the following problem. We have two actions:

- Action $a$: Precondition $P$; effect $Q$

- Action $b$: Precondition $\neg Q$; effect $\neg P$

Actions $a$ and $b$ have no nonconcurrency constraints, thus they are not required to be concurrent to have their specified effects when considered in isolation. Suppose our initial state is $\{P, \neg Q\}$ and the goal state is $\{\neg P, Q\}$. The only plan that achieves this goal requires that $a$ and $b$ be executed concurrently. If we order one before the other, we will destroy the ability to perform the second, and the goal will not be reachable. Thus, POMP can solve this problem while POP could not solve the ESA problem (as formulated above).

To deal with the more general case, we extend the construction of the ESA problem by including (in addition to the actions used in the restricted case) a joint action in the ESA problem for any set of actions $A$ satisfying the following conditions:

- Each element of $A$ is permitted to be executed concurrently (but need not be forced to be concurrent).

- Each element of $A$ clobbers the precondition of some other element of $A$.

- No element of $A$ can be removed without destroying this property.

In other words, we create a joint action corresponding to the concurrent execution of each element of such a set $A$. We'll call these "self-clobbering" joint actions. It should be evident that a concurrent nonlinear plan exists for an arbitrary multiagent planning problem iff there exists a sequence of joint actions (allowing self-clobbering actions) that solve the problem, and hence (by the soundness and completeness of POP) iff POP can find a plan for this generalized ESA problem. We have already seen that POMP can emulate any step of POP involving actions *other than* self-clobbering actions. We simply have to show that POMP can emulate POP's introduction of self-clobbering actions to show completeness.

---

18. This additional flexibility impacts only the soundness of POMP (and is addressed above). In fact, we could have used the current line of reasoning as part of an integrated soundness and completeness proof based on the POP/POMP correspondence, in which case, we would need to explain why this last point does not hinder the soundness of POMP.

Let $A$ be some self-clobbering joint action. We claim that POP is complete (for the generalized ESA problem) if it only ever considers adding $A$ to an incomplete plan when *each* of its elements $a_i \in A$ has an effect that satisfies some subgoal on the agenda. Suppose, to the contrary, that $a_i \in A$ has no consumer on the current agenda. Then either $A$ is not necessary in a successful plan (since some subset of the actions in $A$ can be used), or the actions that consume the effects of some $a_i$ have not yet been introduced. We can discount the former case by considering only executions of POP that do not use this action. POP will be complete even if this action is never considered, since it is able to introduce the individual components (or concurrent subsets) of $A$ that do produce the necessary effects. We can discount the latter case, since there must be a valid execution of POP that introduces the (ultimate) consumers of each element of $a_i$ before introducing $A$. Thus, without loss of generality, we assume that each element $a_i \in A$ satisfies some subgoal on the agenda if $A$ is introduced by POP.

Now suppose POP introduces a self-clobbering action $A$. Since all $a_i \in A$ satisfy some agenda item, POMP can simulate this step as follows: introduce each $a_i$ in turn to satisfy some agenda item, postponing threat resolution among the $a_i$; resolve the self-threats among the $a_i$ through *weak promotion* in the **Causal Link Protection** step (so that we impose ordering constraint $a_i \geq a_j$ for $a_i$ that threatens $a_j$). In the example above, for instance, once actions $a$ and $b$ are added to achieve subgoals $Q$ and $\neg P$, respectively, the only way to resolve the mutual threat is by weak promotion of *both* actions; that is, we impose $a \geq b$ and $b \geq a$. In other words, they are forced to be concurrent. Thus any introduction of a self-clobbering joint action by POP (under the assumptions stated above) has a strong correspondence with a sequence of possible steps in POMP. Since POP can always find a plan under these assumptions, so can POMP. Thus the completeness of POMP in the general case of arbitrary multiagent planning problems is demonstrated.

## 6. Concluding Remarks

One often finds assertions in the planning literature that planning with interacting actions is an inherently problematic affair, requiring substantial extension to existing single-agent planning representations and algorithms. Thus, it is somewhat surprising that only minor changes are needed to enable the STRIPS action representation language to capture interacting actions, and that relatively small modifications to existing nonlinear planners are required to generate concurrent plans. Our solution involves the addition of a concurrent action list to the standard action description, specifying which actions should or should not be scheduled concurrently with the current action in order to achieve a desired effect. The POP planner is augmented by two steps: one which handles the insertion of required concurrent actions, and one which handles threats emanating from the potential concurrent execution of two interfering actions. In addition, explicit reasoning with equality and inequality constraints is introduced. Because of the strong resemblance between our solution for the multiagent case and the solution for the single agent case, little overhead is incurred when actions do not interact. In fact, in the extreme case of non-interacting actions, both our extension to STRIPS and to POP reduce to their single-agent equivalents.

There is a close connection between our specification method and Knoblock's (1994) approach to generating *parallel execution plans*. Knoblock adds to the action description a list that describes the resources used by the action: actions that require the same resource (e.g., access to a database) cannot be scheduled at the same time. Hence, Knoblock's resource list actually characterizes one form of nonconcurrency constraint.[19] In fact, we believe that certain nonconcurrency constraints are more naturally described using such a resource list than with the general method proposed here—augmenting our language with such lists should not prove difficult.

The treatment of concurrent actions in the specification languages $A_c$ (Baral & Gelfond, 1997) and $C$ (Giunchiglia & Lifschitz, 1998) has many features in common with our extension of STRIPS (although $C$, in particular, is a very expressive language with many additional features). These languages allow the use of complex actions—which are sets of primitive actions—analogous to the ability we provide to combine a number of elements into a joint action. Typically, complex actions inherit their effects from the primitive actions contained in them. However, explicit specification of the effects of complex actions is possible, overriding this inheritance. This overriding mechanism can extend to an arbitrary number of levels (e.g., an action $a$ can have some effect, which is overridden when $a$ and $b$ are performed concurrently, but this effect is itself overridden when $c$ is performed as well, etc.). In these action description languages, an implicit view of time is adopted, much like in our treatment, and concurrent actions are assumed to be performed simultaneously. Until quite recently, there were no tools for actually synthesizing plans for domains described in languages such as $C$. However, recent progress in model-based techniques had led to a number of new algorithms, including a SAT encoding for the language $C$ (Giunchiglia, 2000).

When the effects of one agent's actions depend on the actions performed by other agents at the same time, action specification becomes a complex task. The STRIPS representation is useful because it admits a relatively simple planning algorithm. However, despite STRIPS's semantic adequacy and its ability, in principle, to represent any set of actions, verifying that a domain description is accurate becomes more difficult when interactions must be taken into account. Consequently, we believe that the use of dynamic Bayes nets, in conjunction with conditional outcome (or probability) trees (Boutilier & Goldszmidt, 1996), can provide a more natural and concise representation of actions in multiagent settings. This specification technique makes clear the influence of different context conditions on an action's effects, and allows one to exploit the *independence* of different effects. While this representation can be used for stochastic domains, dynamic Bayes nets offer these advantages even in the case of purely deterministic actions. The POMP algorithm naturally extends to this form of domain description, and a more complete treatment of this issue would be an interesting direction for future research.

While adapting existing nonlinear planners to handle interacting actions is conceptually simple, we expect that the increase in domain complexity will inevitably lead to poor computational performance. Indeed, in our experiments with the POMP algorithm, we have found that performance is greatly affected by the ordering of agenda items. Hence, adequate heuristics for making the various choices the planner is faced with—namely, choosing sub-goals, choosing actions that achieve them, and choosing threat-resolution strategies—will become even more critical. Of course, the same issues are central for single-agent nonlinear planners, though we anticipate that the multiagent case with its interacting actions will require different, or additional, heuristics.

---

19. In principle, any nonconcurrency constraint can be handled in this manner by introducing fictitious resources.

An interesting topic for future work would be extending newer planning algorithms such as Graphplan (Blum & Furst, 1995) to handle our multiagent representation language. Indeed, the model-based algorithm of Cimatti, *et al.* (1997) seems to offer promising developments in this direction. Naturally, all representational issues raised in this paper arise regardless of the particular planning algorithm used, although with different implications. For example, the question of whether or not to allow for concurrent actions that destroy one another's preconditions affected which threat removal operators were valid in POMP, whereas in Graphplan they would affect the definition of interfering actions (and consequently, the question of which actions are considered mutually exclusive).

Finally, we note that the approach we have considered is suitable for a team of agents with a common set of goals. It assumes that some central entity generates the plan, and that the agents have access to a global clock or some other synchronization mechanism (this is typically the case for a single agent with multiple effectors, and applies in certain cases to more truly distributed systems). An important research issue is how such plans can be generated and executed in a distributed fashion, and how their execution should be coordinated and controlled. This is an important question to which some answers have emerged in the DAI literature (des Jardins, Durfee, Ortiz Jr., & Wolverton, 1999; Grosz, Hunsberger, & Kraus, 1999; des Jardins & Wolverton, 1999; Boutilier, 1996, 1999; Brafman, Halpern, & Shoham, 1998) and the distributed systems literature (Fagin, Halpern, Moses, & Vardi, 1995).

## Acknowledgments

Thanks to the referees for their suggestions on the presentation of these ideas and to Mike Wellman for his helpful comments. We also thank Daniel Fogel, Ishay Levy, and Igor Razgon for their implementation of the POMP algorithm. Boutilier was supported by NSERC Research Grant OGP0121843, and the NCE IRIS-II program Project IC-7. Brafman was supported by Paul Ivanier Center for Robotics and NCE IRIS-II program Project IC-7. Much of this work was undertaken while both authors were at the University of British Columbia, Department of Computer Science. Preliminary results in this paper were presented in "Planning with Concurrent Interacting Actions," *Proceedings of the Fourteenth National Conference on Artificial Intelligence (AAAI-97)*, Providence, RI, pp.720–729 (1997).

## References

Baral, C., & Gelfond, M. (1997). Reasoning about effects of concurrent actions. *Journal of Logic Programming*, 85–117.

Blum, A. L., & Furst, M. L. (1995). Fast planning through graph analysis. In *Proceedings of the Fourteenth International Joint Conference on Artificial Intelligence*, pp. 1636–1642 Montreal.

Boutilier, C. (1996). Planning, learning and coordination in multiagent decision processes. In *Proceedings of the Sixth Conference on Theoretical Aspects of Rationality and Knowledge*, pp. 195–210 Amsterdam.

Boutilier, C. (1999). Sequential optimality and coordination in multiagent systems. In *Proceedings of the Sixteenth International Joint Conference on Artificial Intelligence*, pp. 478–485 Stockholm.

Boutilier, C., & Goldszmidt, M. (1996). The frame problem and Bayesian network action representations. In *Proceedings of the Eleventh Biennial Canadian Conference on Artificial Intelligence*, pp. 69–83 Toronto.

Brafman, R. I., Halpern, J. Y., & Shoham, Y. (1998). On the knowledge requirements of tasks. *Artificial Intelligence, 98*(1-2), 317–350.

Cimatti, A., Giunchiglia, E., Giunchiglia, F., & Traverso, P. (1997). Planning via model checking: A decision procedure for AR. In *Proceedings of the Fourth European Conference on Planning (ECP'97)*, pp. 130–142 Toulouse.

de Giacomo, G., Lésperance, Y., & Levesque, H. J. (1997). Reasoning about concurrent execution, prioritized interrupts, and exogenous actions in the situation calculus. In *Proceedings of the Fifteenth International Joint Conference on Artificial Intelligence*, pp. 1221–1226 Nagoya.

Dean, T., & Kanazawa, K. (1989). Persistence and probabilistic projection. *IEEE Trans. on Systems, Man and Cybernetics, 19*(3), 574–585.

des Jardins, M. E., Durfee, E. H., Ortiz Jr., C. L., & Wolverton, M. J. (1999). A survey of research in distributed continual planning. *AI Magazine, 20*(4), 13–22.

des Jardins, M. E., & Wolverton, M. J. (1999). Coodinating a distributed planning system. *AI Magazine, 20*(4), 13–22.

Donald, B. R., Jennings, J., & Rus, D. (1993). Information invariants for cooperating autonomous mobile robots. In *Proceedings of the International Symposium on Robotics Research* Hidden Valley, PA.

Durfee, E. H., & Lesser, V. R. (1989). Negotiating task decomposition and allocation using partial global planning. In Huhns, M., & Gasser, L. (Eds.), *Distributed AI*, Vol. 2. Morgan Kaufmann.

Durfee, E. H., & Lesser, V. R. (1991). Partial global planning: A coordination framework for distributed hypothesis formation. *IEEE Transactions on System, Man, and Cybernetics, 21*(5), 1167–1183.

Ephrati, E., Pollack, M. E., & Rosenschein, J. S. (1995). A tractable heuristic that maximizes global utility through plan combination. In *Proceedings of the First International Conference on Multiagent Systems*, pp. 94–101 San Francisco.

Fagin, R., Halpern, J. Y., Moses, Y., & Vardi, M. Y. (1995). *Reasoning about Knowledge*. MIT Press, Cambridge, MA.

Fikes, R., & Nilsson, N. (1971). STRIPS: A new approach to the application of theorem proving to problem solving. *Artificial Intelligence*, *2*(3–4), 189–208.

Ghallab, M., & Alaoui, A. M. (1989). Managing efficiently temporal relations through indexed spanning trees. In *Proceedings of the Eleventh International Joint Conference on Artificial Intelligence*, pp. 1297–1303 Detroit.

Giunchiglia, E. (2000). Planning as satisfiability with expressive action languages: Concurrency, constraints and nondeterminism. In *Proceedings of the Seventh International Conference on Principles of Knowledge Representation and Reasoning (KR'00)*, pp. 657–666 Breckenridge, CO.

Giunchiglia, E., & Lifschitz, V. (1998). An action language based on causal explanation: Preliminary report. In *Proceedings of the Fifteenth National Conference on Artificial Intelligence*, pp. 623–630 Madison, WI.

Grosz, B. J., Hunsberger, L., & Kraus, S. (1999). Planning and acting together. *AI Magazine*, *20*(4), 13–22.

Jensen, R. M., & Veloso, M. M. (2000). OBDD-based universal planning for synchronized agents in non-deterministic domains. *Journal of Artificial Intelligence Research*, *13*, 189–226.

Kautz, H., & Selman, B. (1996). Pushing the envelope: Planning, propositional logic, and stochastic search. In *Proceedings of the Thirteenth National Conference on Artificial Intelligence*, pp. 1194–1201 Portland, OR.

Khatib, O., Yokoi, K., Chang, K., Ruspini, D., Holmberg, R., Casal, A., & Baader, A. (1996). Force strategies for cooperative tasks in multiple mobile manipulation systems. In Giralt, G., & Hirzinger, G. (Eds.), *Robotics Research 7, The Seventh International Symposium*, pp. 333–342. Springer-Verlag, Berlin.

Knoblock, C. A. (1994). Generating parallel execution plans with a partial-order planner. In *Proceedings of the Second International Conference on AI Planning Systems*, pp. 98–103 Chicago.

Koehler, J. (1998). Planning under resource constraints. In *Proceedings of the Thirteenth European Conference on Artificial Intelligence*, pp. 489–493 Brighton, UK.

Lin, F., & Shoham, Y. (1992). Concurrent actions in the situation calculus. In *Proceedings of the Tenth National Conference on Artificial Intelligence*, pp. 590–595 San Jose.

Lingard, A. R., & Richards, E. B. (1998). Planning parallel actions. *Artificial Intelligence*, *99*(2), 261–324.

McCarthy, J., & Hayes, P. (1969). Some philosophical problems from the standpoint of artificial intelligence. *Machine Intelligence*, *4*, 463–502.

Moses, Y., & Tennenholtz, M. (1995). Multi-entity models. *Machine Intelligence, 14*, 63–88.

Penberthy, J. S., & Weld, D. S. (1992). UCPOP: A sound, complete, partial order planner for ADL. In *Proceedings of the Third International Conference on Principles of Knowledge Representation and Reasoning (KR'92)*, pp. 103–114 Cambridge, MA.

Pinto, J. (1998). Concurrent actions and interacting effects. In *Proceedings of the Sixth International Conference on Principles of Knowledge Rerpresentation and Reasoning (KR'98)*, pp. 292–303 Trento.

Reiter, R. (1978). On closed world databases. In Gallaire, H., & Minker, J. (Eds.), *Logic and Databases*, pp. 55–76. Plenum, New York.

Reiter, R. (1991). The frame problem in the situation calculus: A simple solution (sometimes) and a completeness result for goal regression. In Lifschitz, V. (Ed.), *Artificial Intelligence and Mathematical Theory of Computation (Papers in Honor of John McCarthy)*, pp. 359–380. Academic Press, San Diego.

Reiter, R. (1996). Natural actions, concurrency and continuous time in the situation calculus. In *Proceedings of the Fifth International Conference on Principles of Knowledge Representation and Reasoning (KR'96)*, pp. 2–13.

Stone, P., & Veloso, M. M. (1999). Task decomposition, dynamic role assignment, and low-bandwidth communication for real-time strategic teamwork. *Artificial Intelligence, 110*(2), 241–273.

Weld, D. S. (1994). An introduction to least commitment planning. *AI Magazine, 15*(4), 27–61.

Wilkins, D. E., & Myers, K. L. (1998). A multiagent planning architecture. In *Proceedings of the Fourth International Conference on AI Planning Systems*, pp. 154–162 Pittsburgh.

Wolverton, M. J., & des Jardins, M. (1998). Controlling communication in distributed planning using irrelevance reasoning. In *Proceedings of the Fifteenth National Conference on Artificial Intelligence*, pp. 868–874 Madison, WI.